\documentclass[lettersize,journal]{IEEEtran}
\usepackage{changes}
\usepackage{amsmath,amsfonts}
\usepackage{algorithmic}
\usepackage{algorithm}
\usepackage{array}
\usepackage{caption}
\usepackage[caption=false,font=normalsize,labelfont=sf,textfont=sf]{subfig}
\usepackage{textcomp}
\usepackage{amssymb}
\usepackage{wasysym}
\usepackage{stfloats}
\usepackage{url}
\usepackage{verbatim}
\usepackage{graphicx}
\usepackage{hyperref}
\usepackage{cite}
\usepackage{makecell}
\usepackage{multirow}
\usepackage{xcolor}
\usepackage{soul}
\usepackage{booktabs}
\usepackage{stfloats}
\usepackage{CJKutf8}
\usepackage{graphicx, makecell, booktabs}
\usepackage[numbers]{natbib}
\usepackage{graphicx}       % for \resizebox
\usepackage{booktabs}       % for \toprule, \midrule, \bottomrule
\usepackage{array}          % for column formatting
\usepackage{multirow}       % for \multirow
\usepackage[table]{xcolor}  % for \cellcolor and column colors
\usepackage{amsmath}   
\usepackage{caption}
\usepackage{float}  % 使用这个包来处理浮动体的精确位置
\usepackage{afterpage}  % 用于强制控制浮动体的显示位置

\begin{document}

\title{MCL-AD: Multimodal Collaboration Learning for Zero-Shot 3D Anomaly Detection}
% , Min~Li, Tianyi~Wang, Jiachen~Li, Delong~Han, Xinyi~Wu, Zhenyao~Wu, Jin~Wan
\author
{Gang~Li,~\IEEEmembership{Member,~IEEE}, Tianjiao~Chen, Mingle~Zhou, Min~Li, Delong~Han, Jin~Wan
~\IEEEmembership{}
~\IEEEmembership{}
~\IEEEmembership{}% <-this % stops a space
\IEEEcompsocitemizethanks{\IEEEcompsocthanksitem This work was supported by Key R\&D Program of Shandong Province, China (2023CXGC010112), the Taishan Scholars Program (NO. tsqn202103097). \textit{(Corresponding author: Jin~Wan)}

Gang~Li, Tianjiao~Chen, Mingle~Zhou, Min~Li, Delong~Han, Jin~Wan are with Key Laboratory of Computing Power Network and Information Security, Ministry of Education, Shandong Computer Science Center (National Supercomputer Center in Jinan), Qilu University of Technology (Shandong Academy of Sciences), Jinan, China 
and also with Shandong Provincial Key Laboratory of Computing Power Internet and Service Computing, Shandong Fundamental Research Center for Computer Science, Jinan, China (e-mail: lig@qlu.edu.cn; 10431230141@qlu.edu.cn; zhouml@qlu.edu.cn; limin@qlu.edu.cn; handl@qlu.edu.cn; wanj@qlu.edu.cn).

% Tianyi~Wang is a Postdoc at the College of Computing and Data Science, Nanyang Technological University (e-mail: terry.ai.wang@gmail.com).

% Xinyi~Wu and Zhenyao~Wu are with Honor Inc. (e-mail: xinyiwu1996@gmail.com; zhenyaowu1996@gmail.com).

}% <-this % stops an unwanted space

% \author{IEEE Publication Technology,~\IEEEmembership{Staff,~IEEE,}
        % <-this % stops a space
\thanks{}% <-this % stops a space
\thanks{}}

% The paper headers
\markboth{Journal of \LaTeX\ Class Files,~Vol.~14, No.~8, August~2021}%
{Shell \MakeLowercase{\textit{et al.}}: A Sample Article Using IEEEtran.cls for IEEE Journals}

\IEEEpubid{}

\maketitle

\begin{abstract}
Zero-shot 3D (ZS-3D) anomaly detection aims to identify defects in 3D objects without relying on labeled training data, making it especially valuable in scenarios constrained by data scarcity, privacy, or high annotation cost. However, most existing methods focus exclusively on point clouds, neglecting the rich semantic cues available from complementary modalities such as RGB images and texts priors. This paper introduces MCL-AD, a novel framework that leverages multimodal collaboration learning across point clouds, RGB images, and texts semantics to achieve superior zero-shot 3D anomaly detection. Specifically, we propose a Multimodal Prompt Learning Mechanism (MPLM) that enhances the intra-modal representation capability and inter-modal collaborative learning by introducing an object-agnostic decoupled text prompt and a multimodal contrastive loss. In addition, a collaborative modulation mechanism (CMM) is proposed to fully leverage the complementary representations of point clouds and RGB images by jointly modulating the RGB image-guided and point cloud-guided branches. Extensive experiments demonstrate that the proposed MCL-AD framework achieves state-of-the-art performance in ZS-3D anomaly detection.
\end{abstract}


\section{Introduction}
\IEEEPARstart{A}{nomaly} detection aims to identify anomalous regions from given samples~\cite{ref1,ref2}, encompassing both classification and segmentation tasks, and has been widely applied in various domains such as industrial inspection~\cite{ref3,ref4,ref12,ref13,ref48}. Although 2D anomaly detection has achieved significant progress in leveraging RGB information~\cite{ref14,ref15,ref18,ref19,ref49}, real-world anomalies often exhibit complex spatial characteristics, which makes it challenging to rely solely on RGB data in scenarios where defects share similar appearances with the background or foreground. Point cloud-based 3D anomaly detection methods, which exploit rich geometric information, have emerged as an important research direction~\cite{ref20}. 

Current 3D anomaly detection methods typically identify anomalies by measuring the distance between the features of test points and the normal point features saved during the training process~\cite{ref24,ref25,ref26}. However, they often rely on object-specific training data, which limits their generalizability to unseen objects. In contrast, Zero-Shot 3D (ZS-3D) anomaly detection eliminates the need for object-specific training data but requires the model to generalize across semantic categories, which poses significant challenges for adaptability and transferability. 

\begin{figure*}
 % \vspace{-0.2cm}
    \centering
    \includegraphics[width=1\linewidth]{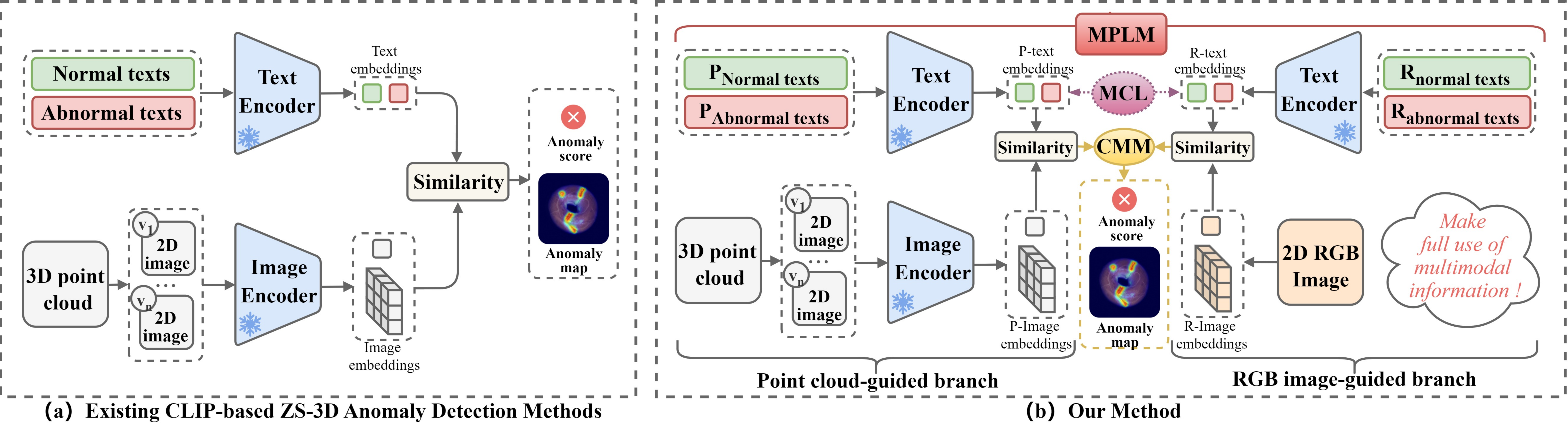}
    \caption{\textrm{Comparison between existing CLIP-based ZS-3D anomaly detection methods and the proposed MCL-AD framework.
(a) Existing methods rely solely on the similarity between 3D data and textual descriptions, without leveraging complementary modalities.
(b) MCL-AD introduces a Multimodal Prompt Learning Mechanism (MPLM) and a Collaborative Modulation Mechanism (CMM) to effectively fuse RGB images, point clouds, and texts semantics, enabling more superior anomaly detection.}}
    \label{fig:1}
    % \vspace{0.1cm}
     % \vspace{-0.1cm}
\end{figure*}

Current ZS-3D anomaly detection methods, particularly those leveraging vision-language models (VLMs) like MVP-PCLIP~\cite{ref27}, have demonstrated promising results. As shown in Fig.\ref{fig:1}(a), these methods primarily focus on 3D modality, such as point clouds, which provide rich geometric information but lack critical texture and color cues. Moreover, they fail to fully capitalize on the complementary strengths of multimodal data, which limits their ability to collaboratively detect anomalies that both appearance and geometry.

In this paper, we propose a ZS-3D anomaly detection framework via multimodal collaboration learning, named MCL-AD, as illustrated in Fig.~\ref{fig:1}(b). MCL-AD achieves superior performance in zero-shot 3D anomaly detection by effectively bridging the gap between point clouds, RGB images, and text semantics, leveraging multimodal collaboration learning to enhance both intra-modal representation and inter-modal interaction. Firstly, we propose a Multimodal Prompt Learning Mechanism (MPLM) that enhances the representational capacity of each modality and facilitates collaborative learning across modalities. Specifically, we design an object-agnostic decoupled text prompt, which explicitly separates the prompts for RGB image and point cloud datas, enabling effective intra-modal feature learning. Furthermore, we propose a Multimodal Contrastive Loss (MCL) that promotes complementary learning between RGB images and point clouds prompts. This loss constructs semantic contrastive relationships by increasing the feature distance between normal and anomalous states, while reducing the distance between RGB and point cloud prompts within the same state. Finally, to mitigate modality imbalance and enhance multimodal fusion during inference, we introduce a Collaborative Modulation Mechanism (CMM). This mechanism dynamically adjusts the outputs of the RGB and point cloud branches to effectively coordinate complementary information across modalities. As a result, the MCL-AD framework demonstrates superior anomaly detection performance in zero-shot 3D scenarios. Evaluated on public datasets including MVTec3D-AD~\cite{ref33} and Eyecandies~\cite{ref33}, our method achieves superior performance compared to existing state-of-the-art methods. 

In summary, our main contributions are as follows:

\begin{itemize}

\item{We propose a novel zero-shot 3D anomaly detection framework, termed MCL-AD, based on multimodal collaboration learning among RGB images, point clouds, and textual semantics. This framework bridges the representational gap between different modalities and enables superior detection without requiring object-specific training data.}

\item{A Multimodal Prompt Learning Mechanism (MPLM) is proposed to enhance the intra-modal representation capability and inter-modal collaborative learning by introducing an object-agnostic decoupled text prompt and a multimodal contrastive loss.}

\item{We propose a Collaborative Modulation Mechanism (CMM) to enhance multimodal fusion and mitigate modality imbalance during inference by jointly modulating the RGB image-guided and point cloud-guided branches through a dual-modality comparison architecture.}

\item{Extensive experiments demonstrate that the proposed MCL-AD outperforms existing methods in zero-shot 3D anomaly detection on the MVTec3D-AD and Eyecandies datasets, while also exhibiting strong cross-dataset generalization.}

\end{itemize}
% \vspace{0.6cm}
\section{Related Work}
\subsection{3D Anomaly Detection}
\label{sec:formatting}
%-------------------------------------------------------------------------

Although 2D image anomaly detection has made significant progress~\cite{ref28,ref8,ref30,ref7,ref32}, 3D anomaly detection methods remain in the early stages of development. Datasets such as MVTec3D-AD~\cite{ref20} and Eyecandies~\cite{ref33}, which contain point clouds paired with corresponding 2D RGB images, have helped bridge the gap between 2D and 3D anomaly detection and have facilitated advances in 3D anomaly detection research. 3D-ST~\cite{ref34} employs a teacher–student network to extract dense local geometric descriptors and performs anomaly detection by matching these descriptors through the student network. AST~\cite{ref35} further improves 3D anomaly detection by introducing an asymmetric teacher–student architecture. IMRNet~\cite{ref36} and 3DSR~\cite{ref37} detect 3D anomalies by computing reconstruction errors. In addition, M3DM~\cite{ref25} aligns 3D features to 2D space and builds separate memory banks for RGB, 3D, and fused features. While effective, this memory construction process incurs high memory usage and computational overhead. To reduce the cost of anomaly detection, EasyNet~\cite{ref19} significantly improves inference speed by introducing an entropy-based attention module with a lightweight reconstruction architecture. CFM~\cite{ref40} proposes an efficient framework that learns to map from one modality to another using nominal samples and detects anomalies by observing discrepancies between the learned mappings. In contrast to the above approaches, we propose a multimodal collaboration learning framework that fully exploits the color and texture information provided by RGB images and the depth and geometric deformation cues captured in point clouds. By collaboratively integrating features from both RGB images and point clouds, our method achieves superior 3D anomaly detection performance.

\subsection{Zero-shot Anomaly Detection}
In recent years, zero-shot anomaly detection methods have gained widespread attention due to their low reliance on labeled data, especially in scenarios where large-scale annotated datasets are unavailable. In the field of 2D image anomaly detection, WinCLIP~\cite{ref9} is the first work applying CLIP to zero-shot image anomaly detection. By designing handcrafted textual prompts, WinCLIP fully leverages CLIP’s strong generalization capabilities for anomaly classification and introduces a window-based fine-grained segmentation strategy to improve segmentation accuracy. Subsequently, April-GAN~\cite{ref41} further enhances anomaly detection performance by utilizing both shallow and deep features from CLIP’s image encoder. SAA~\cite{ref42} integrates multiple VLMs with hybrid prompt regularization to improve detection accuracy across various anomaly detection tasks. Compared to zero-shot 2D, Zero-Shot 3D (ZS-3D) anomaly detection faces greater challenges due to the higher cost and difficulty of collecting and annotating 3D normal data. To address these issues, MVP-PCLIP~\cite{ref27} projects point cloud data into multi-view depth images and leverages a pretrained VLM to perform zero-shot anomaly detection on point clouds. However, existing ZS-3D methods typically focus on a single modality and fail to fully exploit the potential of multimodal data. To overcome this limitation, we propose a multimodal learning framework that integrates point clouds, RGB images and textual modalities to enhance both the accuracy and generalization of ZS-3D anomaly detection.
\begin{figure*}
 % \vspace{-0.2cm}
    \centering
    \includegraphics[width=1\linewidth]{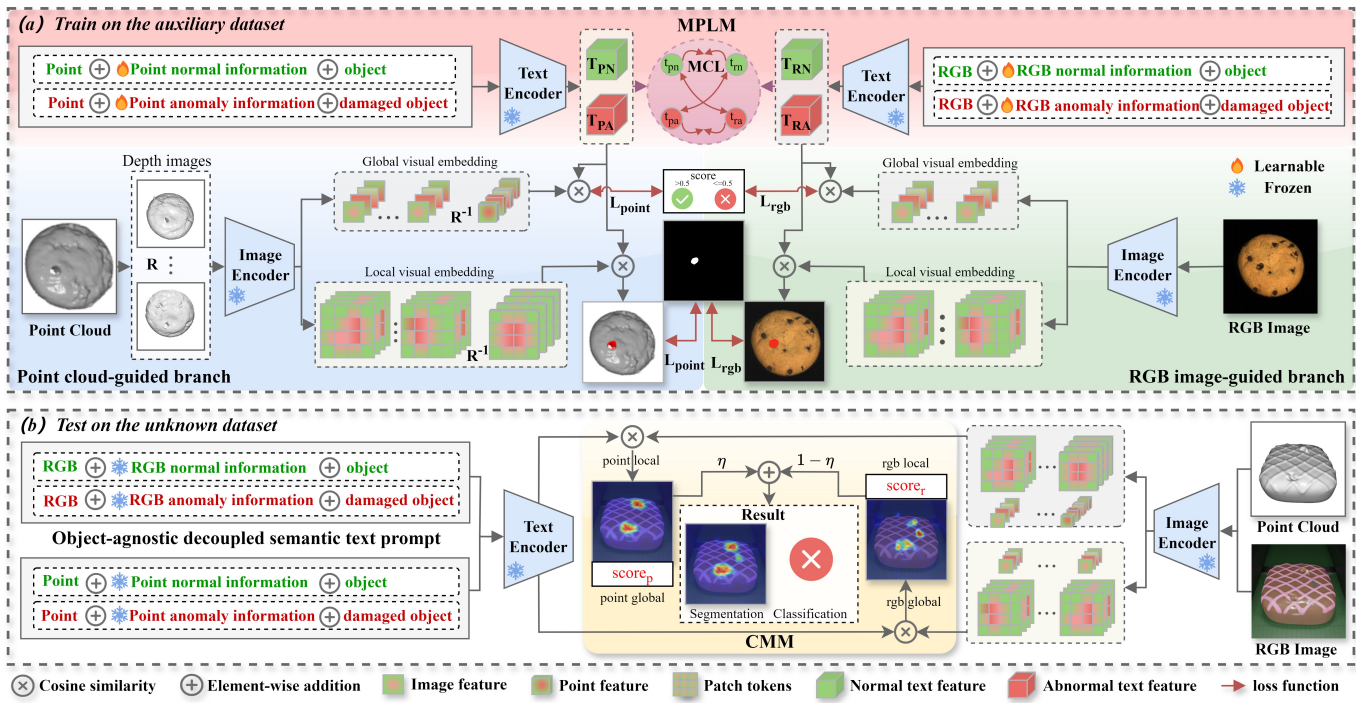}
    \caption{\textrm{
    % The framework of the proposed MCL-AD. (a) The training process of the proposed framework. (b) The test process, which dynamically modulates the prediction results between two branches.
    The framework of the proposed MCL-AD. 
(a) During training, the Multimodal Prompt Learning Mechanism (MPLM) is employed to jointly learn anomaly patterns in point clouds and RGB images using object-agnostic decoupled text prompts, with the multimodal contrastive loss further enhancing the complementary learning between RGB images and point clouds.
(b) During testing, the Collaborative Modulation Mechanism (CMM) dynamically adjusts the prediction results by jointly modulating the RGB image-guided and point cloud-guided branches, fully leveraging their complementary representations to improve accuracy.
    }}
    \label{fig:2}
     % \vspace{-0.1cm}
\end{figure*}
\subsection{Prompt Learning}
The core idea of prompt learning lies in designing appropriate textual or visual prompts to guide pretrained models in adapting to downstream tasks. This approach shows significant performance improvements, particularly in low-data regimes. Anomaly detection often faces the challenge of having abundant normal samples but scarce anomalous instances. As a result, prompt learning sees wide adoption in anomaly detection, especially under zero-shot settings, due to its strong generalization capability in the absence of labeled anomalies. CLIP~\cite{ref29} demonstrates remarkable zero-shot learning ability and finds applications in various tasks, including image classification and object detection. WinCLIP~\cite{ref9} introduces CLIP into anomaly classification and segmentation tasks, and enhances CLIP’s generalization performance by designing a compositional prompt set. Building on this, CoOp~\cite{ref38} and MaPLe~\cite{ref39} propose learnable prompt mechanisms that automatically optimize prompts for specific tasks, further improving model performance. In addition, AnomalyCLIP~\cite{ref46} introduces an object-agnostic prompt strategy, which allows CLIP to adapt to anomaly detection without relying on predefined category priors. In contrast to these methods, we propose an object-independent decoupled semantic prompt within the MCL-AD framework, which enables each modality to independently optimize its feature representation through the explicit separation of RGB image and point cloud prompts. Moreover, we introduce a multimodal contrastive loss that encourages inter-modal synergy and complementarity, thereby enhancing the  accuracy of the model in ZS-3D anomaly detection.

\section{Method}

\subsection{Approach overview}
MCL-AD is proposed to perform zero-shot 3D (ZS-3D) anomaly detection by effectively leveraging multimodal information from point clouds, RGB images, and text prompts. First, both RGB images and depth views rendered from point clouds are processed through frozen CLIP image encoders to extract rich visual features from each modality (Sec.~\ref{sec:3.2}). As illustrated in Fig.\ref{fig:2}(a), during the training phase, we introduce a Multimodal Prompt Learning Mechanism (MPLM) to enhance the representational power of each modality. Specifically, MPLM incorporates an Object-Agnostic Decoupled Text Prompt (ODTP) that separately models textual prompts for the RGB and point cloud branches. To further optimize the prompt, we propose a Multimodal Contrastive Loss (MCL), which enforces semantic consistency between normal and anomalous prompts across modalities (Sec.~\ref{sec:3.3}). During training, the model is jointly optimized with the RGB loss, point cloud loss, and multimodal contrastive loss. (Sec.~\ref{sec:3.4}). As illustrated in Fig.\ref{fig:2}(b), during the test phase, we propose a Collaborative Modulation Mechanism (CMM) to facilitate cross-modal representation by adaptively calibrating the outputs of the RGB and point cloud branches (Sec.\ref{sec:3.5}).
Through this multimodal collaboration learning, MCL-AD achieves robust generalization to novel objects and complex anomalies, even without access to object-specific training data.

\subsection{Point Clouds and RGB Images Feature Representations}
\label{sec:3.2}
Effective acquisition of point cloud and RGB image features is essential for enabling both intra-modal learning and inter-modal collaborative representation. First, a multi-view rendering technique is employed to convert sparse 3D point clouds into dense 2D depth images, effectively narrowing the modality gap and enabling CLIP-based feature extraction. Next, we independently extract and organize global and local features from both RGB images and the rendered depth views using a frozen CLIP encoder. Finally, an inverse rendering strategy is introduced to map the extracted 2D features back into the 3D point cloud space, constructing rich and aligned representations across both modalities to facilitate collaborative anomaly detection.

\textbf{Multi-View Images Generation.} MCL-AD involves processing two different visual modalities: 2D images and 3D point clouds. Unlike 2D images, 3D point clouds are inherently sparse and irregular, leading to significant domain discrepancies when processed directly with CLIP. To address this issue, we adopt a multi-view rendering approach that projects 3D point cloud data into multiple 2D depth images from different viewpoints~\cite{ref43}.

Specifically, given a point cloud dataset defined as $D_{3d} = \{(P_{3d,i}, M_{3d,i})\}_{i=1}^N$, we define a set of rendering matrices $R(k)$ for each viewpoint, where $k \in K$ denotes the index of the viewpoint and $K$ is the total number of views. Each rendering matrix
$R(k)$ represents a rotation along the x-axes and y-axes by angles $\theta_x$ and $\theta_y$, respectively, defined as:
\begin{equation} 
\begin{split}
\label{e29}
R(k)(\theta_x) = \begin{bmatrix}
1 & 0 & 0 \\
0 & \cos(\theta_x) & -\sin(\theta_x) \\
0 & \sin(\theta_x) & \cos(\theta_x)
\end{bmatrix},
\end{split}
\end{equation}

\begin{equation} 
\begin{split}
\label{e29}
R(k)(\theta_y) = \begin{bmatrix}
\cos(\theta_y) & 0 & \sin(\theta_y) \\
0 & 1 & 0 \\
-\sin(\theta_y) & 0 & \cos(\theta_y)
\end{bmatrix}.
\end{split}
\end{equation}

Using these transformations, we render both the point clouds and the corresponding point-level ground truth from different viewpoints, resulting in 2D projections and their associated labels. Specifically, for each sample $i$ and view $k$, the rendered outputs are:
% \vspace{-0.2cm}
\begin{equation} 
\begin{split}
\label{e29}
P^{(k)}_i = R(k)(P_{3d,i}),
\end{split}
\end{equation}
% \vspace{-0.5cm}
\begin{equation} 
\begin{split}
\label{e29}
M^{(k)}_i = R(k)(M_{3d,i}),
\end{split}
\end{equation}
where $P^{(k)}_i \in \mathbb{R}^{H \times W}$ denotes the 2D depth image rendered from the $k_{th}$ view, and  $M^{(k)}_i \in \mathbb{R}^{H \times W}$ is the corresponding 2D ground truth mask. Anomalous pixels are labeled as 1, while normal pixels are labeled as 0. Additionally, the global label for the entire image is denoted by $G^{(k)}_i \in [0,1]$. By this method, we can generate high-precision multi-view images from different perspectives, which is helpful for better capturing details and conducting anomaly detection.

\textbf{Representations for Point Cloud and RGB Information.} Point clouds and RGB images are the two most commonly used modalities in 3D anomaly detection. RGB images provide rich texture information, while point clouds capture the geometric shape and spatial location of objects. Although each modality offers distinct advantages, relying on a single modality for anomaly detection is inherently limited. Therefore, it is essential to learn independent feature representations for each modality to better capture anomaly-specific characteristics. Within the MCL-AD framework, we extract features from both point clouds and RGB images data through a series of steps and project them into a unified representation space.

RGB Image Features. During the training phase, we utilize a frozen CLIP image encoder to extract feature representations from RGB images. By feeding an RGB image $I_{\text{rgb}}$ into the CLIP encoder, we obtain its global feature $f_{\text{rgb}}^{\text{global}} \in \mathbb{R}^D$ and local feature $f_{\text{rgb}}^{\text{local(m)}} \in \mathbb{R}^D$ (m denotes the index of the feature map layer, where $m \in [0, 1, 2, 3]$). These features capture both global structural information and local detailed patterns of the image, which are beneficial for subsequent anomaly detection tasks. The extracted RGB image features can be represented as:
\begin{equation} 
\begin{split}
\label{e29}
f_{rgb}=CLIP_{image}(I_{rgb})=[f_{\text{rgb}}^{\text{global}},f_{\text{rgb}}^{\text{local(m)}}].
\end{split}
\end{equation}

Point Cloud Features. For point cloud data $D_{\text{3d}}$, we first render the 3D point cloud from multiple views to obtain a series of 2D depth images. Each depth image $P^{(k)}_i$ corresponds to the point cloud projected from the $k_{th}$ view. Using the CLIP image encoder, we extract the global and local features$f_i^{global(k)}$ and $f_i^{local(k)(m)}$ from each depth image. The feature extraction process for the depth images is described as:
\begin{equation} 
\begin{split}
\label{e29}
f_{PC}=CLIP_{image}(P^{(k)}_i)=[f_i^{global(k)} ,f_i^{local(k)(m)} ].
\end{split}
\end{equation}

To obtain the global feature $p_i^{global}$ of the point cloud, we divide $f_i^{global(k)} $ into feature blocks corresponding to different viewpoints and compute their weighted average along the dimension $dim=1$:
\begin{equation} 
\begin{split}
\label{e29}
p_i^{global} = \text{mean} (\text{stack} (\text{chunk}(f_i^{global(k)}, nv, \text{dim} = 0),\\ \text{dim} = 1),
\text{dim} = 1),
\end{split}
\end{equation}
where $nv$ denotes the number of viewpoints. We then apply inverse rendering to integrate features from the depth images back into the point cloud space, obtaining the local feature $f_i^{global(k)} $ as:
\begin{equation} 
\begin{split}
\label{e29}
p_i^{local(m)} =InverseRender(f_i^{local(k)(m)} ),
\text{dim} = 1),
\end{split}
\end{equation}
where $\text{chunk}(\cdot)$ denotes splitting the image features into $nv$ parts corresponding to different viewpoints, and $\text{stack}(\cdot)$ aggregates them along the viewpoint dimension. $\text{mean}(\cdot)$ computes the average, while $\text{InverseRender}(\cdot)$ refers to the inverse rendering process, which maps CLIP-extracted features from image space back to the corresponding global and local point cloud space.

Finally, the overall point cloud feature is given by:
\begin{equation} 
\begin{split}
\label{e29}
p_i = \left[ p_i^{\text{global} }, p_i^{\text{local}(m) } \right].
\end{split}
\end{equation}

\subsection{Multimodal Prompt Learning Mechanism}
\label{sec:3.3}
In RGB images, color information is crucial for effectively identifying texture anomalies, while depth information in point clouds plays a key role in detecting geometric deformations in 3D space. By independently configuring corresponding textual prompts and leveraging the relationships between these prompts, the synergistic effects across modalities can be further enhanced. Specifically, we propose an object-agnostic decoupled text prompt that separates the feature learning processes for RGB and point cloud data, while generating class-agnostic prompts to mitigate category bias. These prompts are carefully designed for both normal and anomalous conditions, allowing the model to learn semantic representations associated with each state. In addition, we introduce a Multimodal Contrastive Loss (MCL) that aligns the text embeddings of different semantic states across RGB and point cloud prompts.  

\textbf{Object-agnostic Decoupled Text Prompt.} RGB image prompts primarily capture global structural features of images, such as texture and color, helping the model identify anomalies at the image level. In contrast, point cloud prompts focus on capturing geometric information of objects, especially local defects or deformations, such as scratches or distortions. To enable the model to learn more fine-grained and specialized features from each modality and to improve its ability to detect subtle differences between normal and anomalous instances, we propose an object-agnostic decoupled text prompt design within the MCL-AD framework. This design independently optimizes the feature learning process for each modality and avoids category bias by generating general and class-agnostic semantic prompts for both RGB images and point cloud data.

We construct learnable text prompts by introducing class-agnostic normal prompts ($t_n$) and anomaly prompts ($t_a$), which capture and represent the object characteristics under normal and anomaly conditions, respectively. The text prompts are expressed in a binary form, following a target-independent prompt structure, as detailed below:
\begin{equation} 
\begin{split}
\label{e29}
t_n = [N_i][object],
\end{split}
\end{equation}
% \vspace{-0.51cm}
\begin{equation} 
\begin{split}
\label{e29}
t_a =[A_j][damaged][object],
\end{split}
\end{equation}
where $[N_i]$ (where $i \in [1,L_n]$) represents the learnable token under normal conditions, encoding the general state of the object, while $[A_i]$ (where $j \in [1,L_a]$) denotes the learnable token under anomaly conditions, indicating specific defects or damages to the object. Unlike traditional category-based semantic learning methods, these prompts are designed to extract features that distinguish between normal and anomaly conditions from RGB images and point clouds, without relying on information specific to any object class. By incorporating the term ``object" into the prompt structure, the model learns class-agnostic representations, enabling better adaptation to cross-category and anomaly detection tasks. In the anomaly prompts, the term ``damaged" is explicitly added to clearly identify objects in an anomaly state. The specific design of the RGB image and point cloud prompts is as follows:
\begin{equation} 
\begin{split}
\label{e29}
R_n = [R_s][N_i^r][object],
\end{split}
\end{equation}
\vspace{-0.51cm}
\begin{equation} 
\begin{split}
\label{e29}
R_a = [R_s][A_i^r][damaged][object],
\end{split}
\end{equation}
\vspace{-0.51cm}
\begin{equation} 
\begin{split}
\label{e29}
P_n = [P_s][N_i^p][object],
\end{split}
\end{equation}
\vspace{-0.51cm}
\begin{equation} 
\begin{split}
\label{e29}
P_a = [P_s][A_i^p][damaged][object],
\end{split}
\end{equation}
where $R_s$ and $P_s$ represent the text flags for RGB and point cloud, respectively, while $R_n$ and $R_a$ represent the RGB image prompts under normal and anomaly conditions, and $P_n$ and $P_a$ correspond to the same conditions for point cloud prompts. The learnable tokens $[N_i^r], [N_i^p], [A_i^r], [A_i^p]$  ($i\in[1,L_n],j\in[1,L_a]$) correspond to the semantic features under different states.

Finally, the four sets of RGB and point cloud text prompts are embedded through a pre-trained CLIP text encoder to generate the corresponding text prompt embeddings: $e_{\text{rgb}}^{\text{normal}}, e_{\text{rgb}}^{\text{anomaly}}, e_{\text{point}}^{\text{normal}}, e_{\text{point}}^{\text{anomaly}}$.
 
\textbf{Multimodal Contrastive Loss.} Visual understanding is inherently hierarchical. Point cloud prompts focus on capturing the overall geometric structure of objects, while RGB prompts complement this by providing detailed texture information, thereby enriching the perception of the object. To enhance the feature collaboration between the two modalities, we design a Multimodal Contrastive Loss (MCL) to align the textual embeddings of different semantic states. This loss constructs semantic contrastive relationships by increasing the feature distance between normal and anomalous states, while simultaneously reducing the distance between point cloud and RGB prompts within the same state. As a result, it improves the model’s discriminative capability in anomaly detection tasks.

Specifically, the MCL function optimizes the relationships within a set of prompt triplets by using the point cloud prompt embedding as the anchor and adjusting its relation to the RGB prompt embeddings. The distance between the anchor and the positive prompt is minimized, while the distance to the negative prompt is maximized. The loss function is defined as follows:
\begin{equation} 
\footnotesize
\begin{split}
\label{e29}
L_{\text{mcl}} = \frac{1}{N} \sum_{i=1}^{N} \left(\max(0, \text{margin} - \| a_i - n_i \|_2)^2   + \| a_i - p_i \|_2^2 \right),
\end{split}
\end{equation}
where $a_i$, $p_i$, and $n_i$ represent the embeddings of the anchor, positive, and negative prompts, respectively. The term $\| \cdot \|_2$ denotes the Euclidean distance, and margin defines the minimum separation required between the anchor and the negative prompt.

During optimization, when the point cloud normal prompt is used as the anchor, the RGB normal prompt is treated as the positive (to be pulled closer), and the RGB anomalous prompt as the negative (to be pushed away). Conversely, when the point cloud anomalous prompt serves as the anchor, the RGB anomalous prompt is treated as the positive, while the RGB normal prompt becomes the negative. This bidirectional alignment mechanism ensures that the learning process is contextualized with respect to both normal and anomalous semantics in the point cloud modality. As a result, the model learns to better discriminate between normal and anomalous features, enhancing the representational capabilities of each modality.

To further boost the model's performance, we integrate the contrastive losses under both normal and anomalous conditions, leading to the final total loss function:
\begin{equation} 
\begin{split}
\label{e29}
L_{mcl}^{total} = L_{mcl}^{normal} + L_{mcl}^{anomaly},
\end{split}
\end{equation}
where $L_{mcl}^{normal}$ focuses on alignment under normal conditions, while  $L_{mcl}^{anomaly}$ operates under anomalous conditions.

\subsection{Loss Functions}
\label{sec:3.4}
The loss functions in MCL-AD encompass  the RGB loss $L_{\text{rgb}}$, point cloud loss $L_{\text{point}}$, and the multimodal contrastive loss $L_{mcl}^{total}$.

\textbf{3D Point Cloud Representation Learning Loss.} To fully capture 3D semantic information, we design two distinct loss functions that target both local and global anomaly semantics in the point cloud data.

For local anomaly modeling, we compute the local features 
$f_i^{local(k)(0)}$ from the depth image at each view $k$, using the features from the $0_{th}$ layer to reduce computational complexity. These local features are then matrix-multiplied with the textual prompt embedding $e_{\text{point}}^{\text{normal/anomaly}}$ , yielding the view-wise local similarity maps:
\begin{equation} 
\begin{split}
\label{e29}
F_{pd}^{local(k)} = f_i^{local(k)(0)} \otimes  e_{\text{point}}^{\text{normal/anomaly}}.
\end{split}
\end{equation}

These 2D similarity maps are then inverse-rendered back into 3D space using the coordinates from each view, generating the point-level local similarity map:
\begin{equation} 
\begin{split}
\label{e29}
F_{p}^{local}= R^{-1} \left (F_{pd}^{local(k)}, nv \right),
\end{split}
\end{equation}
where $R^{-1}(\cdot)$ denotes the inverse rendering operation from 2D to 3D space, and nv is the number of views.

To precisely model the decision boundary of anomalous regions and alleviate class imbalance, we adopt a combination of Focal Loss and Dice Loss to supervise the local loss component. The local loss function is formulated as:
\begin{equation} 
\footnotesize
\begin{split}
\label{e29}
L_{p}^{local} = \frac{1}{K} \sum_{k} \text{Focal}(F_{pd}^{local(k)}(n) \otimes F_{pd}^{local(k)}(a), M^{(k)}_i ) +\\ \text{Dice}(F_{pd}^{local(k)}(n), I - M^{(k)}_i) + \text{Dice}(F_{pd}^{local(k)}(a), M^{(k)}_0 ) +\\ \text{Dice}(F_{p}^{local}(n), I - M_i) + \text{Dice}(F_{p}^{local}(a), M_i),
\end{split}
\end{equation}
where $M^{(k)}_i$ and $M_i$ denote the ground truth masks for each 2D view and 3D point cloud, respectively. The symbol $\oplus$ represents the concatenation operation, and $I$ denotes an all-one matrix of the same size as $M_i$ or $M^{(k)}_i$.

For global anomaly modeling, we compute global features 
$f_i^{global(k)}$ from the depth image of each view k, and perform matrix multiplication with the textual prompt embedding 
$e_{\text{point}}^{\text{normal/anomaly}}$ to obtain view-wise global similarity maps:
\begin{equation} 
\begin{split}
\label{e29}
F_{pd}^{global(k)} = f_i^{global(k)} \otimes   e_{\text{point}}^{\text{normal/anomaly}}.
\end{split}
\end{equation}

These view-wise similarity maps are then averaged across all views to produce a global anomaly score for the 3D point cloud:
\begin{equation} 
\begin{split}
\label{e29}
F_{p}^{global}= \frac{1}{nv} \sum_{k=1}^{nv} F_{pd}^{global(k)}.
\end{split}
\end{equation}

To supervise the global anomaly detection, we employ cross-entropy loss based on cosine similarity between the similarity maps and their corresponding ground truth masks. The global loss is formulated as:
\begin{equation} 
\begin{split}
\label{e29}
L_{p}^{global} = \frac{1}{N} \sum_{i} \text{CrossEntropy}(F_{pd}^{global(k)}, G^{(k)}_i) +\\ \text{CrossEntropy}(F_{p}^{global}, G_i),
\end{split}
\end{equation}
where $G^{(k)}_i$ denotes the ground truth annotation for each 2D view, and $G_i$ is the ground truth mask for the entire 3D point cloud.
Finally, the overall point cloud loss is defined as the combination of the local and global components:
\begin{equation} 
\begin{split}
\label{e29}
L_{\text{point}} = F_{p}^{local} + L_{p}^{global}.
\end{split}
\end{equation}

\textbf{2D RGB Representation Learning Loss.} Similar to the point cloud branch, we capture semantic information from RGB images from both local and global perspectives. For local anomaly modeling in RGB images, we extract local features $f_{rgb}^{local(i)}$ for each image and compute the local similarity maps by performing matrix multiplication with the textual prompt embedding $e_{\text{rgb}}^{\text{normal/anomaly}}$: 
\begin{equation} 
\begin{split}
\label{e29}
F_{rgb}^{local(i)} = f_{rgb}^{local(i)} \otimes   e_{\text{rgb}}^{\text{normal/anomaly}}.
\end{split}
\end{equation}

To model fine-grained anomalies and address the issue of class imbalance, we adopt a combination of Focal Loss and Dice Loss for supervision. The local loss is defined as:
\begin{equation} 
\small
\begin{split}
\label{e29}
L_{r}^{local} = \frac{1}{N} \sum_{i} \text{Focal}(F_{rgb}^{local(i)}(n) \otimes F_{rgb}^{local(i)}(a), M_i) +\\ \text{Dice}(F_{rgb}^{local(i)}(n), I - M_i) + \text{Dice}(F_{rgb}^{local(i)}(a), M_i) ,
\end{split}
\end{equation}
where $M_i$ denotes the ground truth mask of image $i$, 
$\otimes$ represents the concatenation operation, and $I$ is an all-one matrix of the same size as $M_i$.

For global semantic modeling, we extract global features $f_{rgb}^{global}$ and compute the global similarity map via matrix multiplication with the prompt embedding:
\begin{equation} 
\begin{split}
\label{e29}
F_{rgb}^{global} = f_{rgb} \otimes   e_{\text{rgb}}^{\text{normal/anomaly}}.
\end{split}
\end{equation}

We then compute the global loss using cross-entropy based on cosine similarity, formulated as:
\begin{equation} 
\begin{split}
\label{e29}
L_{rgb}^{global} = \text{CrossEntropy}(F_{rgb}^{global}, G_i),
\end{split}
\end{equation}
where $G_i$ is the ground truth label for the entire RGB image. The total loss for RGB image representation learning is defined as the sum of local and global losses:
\begin{equation} 
\begin{split}
\label{e29}
L_{\text{rgb}} = F_{rgb}^{local} + L_{rgb}^{global}.
\end{split}
\end{equation}

The total loss of the MCL-AD framework is defined as:
\begin{equation} 
\begin{split}
\label{e29}
L_{all} = \lambda_1 L_{\text{point}} + \lambda_2 L_{\text{rgb}} + \lambda_3 L_{mcl}^{total},
\end{split}
\end{equation}
where $\lambda_1$ and $\lambda_2$ are empirically set to 1, while $\lambda_3$ is set to 0.8 based on the experiment in Section~\ref{sec:4.5}.

\subsection{Collaborative Modulation Mechanism}
\label{sec:3.5}
 As illustrated in Fig.~\ref{fig:2}(b), during the inference phase, we employ a dual-modality multi-view comparison architecture comprising an RGB image feature-guided branch and a point cloud feature-guided branch. The RGB image feature-guided branch is responsible for extracting texture and color information, while the point cloud feature-guided branch captures the geometric structure and spatial characteristics of the object. Simultaneously, we CMM to efficiently fuse multimodal information by effectively calibrating the outputs of these two branches.

In the RGB image feature-guided branch, we first extract features from the input RGB image $I_{rgb}
$. Using a pre-trained visual encoder, we obtain the global feature $f_{\text{rgb}}^{\text{global}}$ and multi-level patch features $f_{\text{rgb}}^{\text{local(i)}}$ of the image. The patch tokens from four stages, denoted as $f_{\text{rgb}}^{\text{local(i)}}$, are matched and fused with the RGB text prompts $e_{\text{rgb}}^{\text{normal}}$ and $e_{\text{rgb}}^{\text{anomaly}}$ (details are provided in Sec.~\ref{sec:3.3}) to generate the anomaly detection map based on the RGB image:
\begin{equation} 
\begin{split}
\label{e29}
M_{\text{rgb}} = (\sum_{i=1}^{layer=4} \text{Softmax} \left( B_i \left( f_{\text{rgb}}^{\text{local(i)}} \odot e_{\text{rgb}}^{\text{anomaly}} \right) \right) + \\(1- \sum_{i=1}^{layer=4} \text{Softmax} \left( B_i \left( f_{\text{rgb}}^{\text{local(i)}} \odot e_{\text{rgb}}^{\text{normal}} \right) \right))/2,
\end{split}
\end{equation}
where $B_i$ represents the weighting matrix at each stage, $\odot$ denotes element-wise dot product, and 
$e_{\text{rgb}}^{\text{anomaly/normal}}$ are the corresponding rgb image text prompt features.

For anomaly classification of the RGB image, the global feature $f_{\text{rgb}}^{\text{global}}$ is compared with the text prompt features. By setting a predefined normal threshold, the system determines whether the image contains anomalies:
\begin{equation} 
\footnotesize
\begin{split}
\label{e29}
\text{Score}_{\text{rgb}} = \max \left( \text{Softmax} \left( (f_{\text{rgb}}^{\text{global}} \odot (e_{\text{rgb}}^{\text{anomaly/normal}})^T)/0.07\right) \right),
\end{split}
\end{equation}
where 0.07 is a scaling factor that adjusts the similarity range, and the Softmax function is applied to obtain a probability distribution over the categories.

In the point cloud feature-guided branch, the input point cloud $D_{\text{3d}}$ is first rendered into multi-view 2D depth images $P^{(k)}_i$. Using the same pre-trained visual encoder, we extract the global feature $f_{\text{point}}^{\text{global}(k)}$ and multi-level patch features $f_{\text{point}}^{\text{local(k)(i)}}$ for each view. The patch features from the first stage under nine viewpoints, denoted as $f_{\text{point}}^{\text{local(k)(0)}}$, are matched with the corresponding point cloud text prompts $e_{\text{point}}^{\text{normal}}, e_{\text{point}}^{\text{anomaly}}$ (details are provided in Sec.~\ref{sec:3.3}), and then mapped back to the 3D space according to the positional information of each viewpoint, yielding the anomaly detection map for the point cloud:
\begin{equation} 
\begin{split}
\label{e29}
M_{\text{point
}}^{(k)} =\text{Softmax} \left( B_i \left( f_{\text{point}}^{\text{local(k)(0)}} \odot e_{\text{point}}^{\text{normal/anomaly}} \right) \right),
\end{split}
\end{equation}
\begin{equation} 
\begin{split}
\label{e29}
M_{\text{point
}} = R^{-1} \left (M_{\text{point
}}^{(k)} , d_{3d} \right),
\end{split}
\end{equation}
where $d_{3d}$ represents the 3D positional information of the point cloud, and $R^{-1}(\cdot)$ denotes the inverse rendering process that maps the anomaly detection results from 2D views back to the 3D space. For point cloud anomaly classification, the global features from multiple views $f_{\text{point}}^{\text{global}(k)}$ are compared with the text prompt features. By setting an appropriate normal threshold, we determine whether the point cloud exhibits anomalies:
\begin{equation} 
\footnotesize
\begin{split}
\label{e29}
\text{Score}_{\text{point}}  = \frac{1}{nv} \sum_{k=1}^{nv}  \max \left( \text{Softmax} \left( (f_{\text{rgb}}^{\text{global}} \odot (e_{\text{point
}}^{\text{anomaly/normal}})^T)/0.07\right) \right).
\end{split}
\end{equation}

Due to domain discrepancies across datasets and the uncertainty of detection targets, the learned textual prompts may not fully generalize to unseen scenarios. To ensure the stability of anomaly detection results, we propose a CMM. Specifically, we introduce a hyperparameter $\eta$ to explicitly balance the contributions from the RGB image-guided branch and the point cloud-guided branch. This strategy not only adapts to the domain gap between the training and target datasets, but also addresses the uncertainty inherent in the detection targets, thereby improving the efficiency of the detection process while maintaining high accuracy. The overall process can be formulated as follows:
\begin{equation} 
\begin{split}
\label{e29}
M_{\text{final}} =( \eta \times  G_{\sigma}(M_{\text{rgb}}) + (2 - \eta) \times G_{\sigma}(M_{\text{point}}))/2,
\end{split}
\end{equation}
\begin{equation} 
\begin{split}
\label{e29}
\text{Score}_{\text{final}} = ( \eta \times Score_{\text{rgb}} + (2 - \eta) \times Score_{\text{point}} )/2 + \\(max(M_{\text{rgb}}) + max(M_{\text{point}}))/2,
\end{split}
\end{equation}
where $G_{\sigma}(\cdot)$ denotes the Gaussian filter, and $\eta$ represents the weighting factor that balances the outputs of the two branches. The variations in this hyperparameter $\eta$ are attributed to differences across datasets, and the implementation details will be further discussed in the ablation study (Sec.~\ref{sec:4.4}).

\begin{table*}[b]
\setlength{\tabcolsep}{2pt}
\renewcommand{\arraystretch}{1.5}
\caption{\textrm{Quantitative results on the MVTec3D-AD dataset (\%).}}
\label{tab:1}
\scriptsize
\centering
\resizebox{\linewidth}{!}{
\resizebox{\textwidth}{!}{
\rmfamily
\begin{tabular}{clccccccccccc}
\toprule[0.9pt]
\multicolumn{1}{l}{}                                                                 & Method             & Bagel        & Cable\_gland                & Carrot       & Cookie       & Dowel        & Foam         & Peach        & Potato       & Rope  &Tire    & Mean                               \\ \hline
& CLIP+R.            & (55.1, 85.9)  & (55.0, 84.1)  & (64.5, 90.1)  & (50.6, 83.1)  & (59.1, 84.6)  & (69.0, 90.7)  & (72.0, 91.3)  & (56.7, 85.5)  & (70.8, 86.0)  & (51.7, 82.9)  & (60.4, 86.4)                        \\
& PoinCLIP V2        & (51.6, 83.7)  & (63.8, 87.6)  & (47.7, 83.5)  & (47.8, 78.0)  & (51.8, 80.5)  & (45.2, 78.5)  & (49.2, 78.7)  & (55.4, 82.9)  & (39.1, 62.4)  & (46.0, 76.9)  & (49.8, 79.3)                       \\
& PointCLIP V2a      & (53.4, 84.4)  & (64.7, 89.1)  & (48.0, 83.4)  & (48.4, 78.4)  & (47.1, 81.1)  & (45.9, 79.0)  & (49.6, 79.2)  & (55.5, 85.9)  & (34.9, 60.5)  & (46.1, 76.9)  & (49.4, 79.8)                       \\
& AnomalyCLIP        & (78.8, 93.5)  & (58.1, 84.0)  & (63.2, 88.7)  & (72.3, 89.1)  & (53.8, 82.5)  & (65.1, 89.8)  & (73.7, 91.1)  & (83.1, 93.7)  & (77.5, 89.1)  & (55.2, 82.5)  & (66.2, 87.6)                       \\
& PointAD            & ({\color[HTML]{FF0000} 98.8}, {\color[HTML]{0000EE} 99.7}) & 
({\color[HTML]{0000EE} 79.9}, {\color[HTML]{0000EE} 94.7}) & 
({\color[HTML]{FF0000} 95.5}, {\color[HTML]{0000EE} 98.9}) & 
({\color[HTML]{0000EE} 86.2}, {\color[HTML]{0000EE} 95.5}) & 
({\color[HTML]{FF0000} 98.5}, {\color[HTML]{0000EE} 90.6}) & 
({\color[HTML]{0000EE} 84.4}, {\color[HTML]{0000EE} 96.1}) & 
({\color[HTML]{FF0000} 96.6}, {\color[HTML]{FF0000} 99.1}) & 
({\color[HTML]{0000EE} 90.7}, {\color[HTML]{0000EE} 97.0}) & 
({\color[HTML]{0000EE} 93.6}, {\color[HTML]{0000EE} 97.3}) & 
({\color[HTML]{FF0000} 74.6}, {\color[HTML]{0000EE} 92.0}) & ({\color[HTML]{0000EE} 86.9}, {\color[HTML]{0000EE} 96.1})
 \\
& MVP-WinCLIP        & (-, -)        & (-, -)        & (-, -)        & (-, -)        & (-, -)        & (-, -)        & (-, )        & (-, -)        & (-, -)        & (-, -)        & (71.0, 89.2)                       \\
& MVP-PCLIP          & (-, -)        & (-, -)        & (-, -)        & (-, -)        & (-, -)        & (-, -)        & (-, -)        & (-, -)        & (-, -)        & (-, -)        & (71.9, 89.6)                       \\ \cline{2-13} 
\multirow{-9}{*}{\begin{tabular}[c]{@{}c@{}}Image-level\\ (AUROC, AP)\end{tabular}}  & MCL-AD            &({\color[HTML]{0000EE} 98.5}, {\color[HTML]{FF0000} 99.7}) & 
({\color[HTML]{FF0000} 80.1}, {\color[HTML]{FF0000} 94.8}) & 
({\color[HTML]{0000EE} 95.1}, {\color[HTML]{FF0000} 99.0}) & 
({\color[HTML]{FF0000} 93.4}, {\color[HTML]{FF0000} 98.1}) & 
({\color[HTML]{0000EE} 72.0}, {\color[HTML]{FF0000} 91.8}) & 
({\color[HTML]{FF0000} 85.0}, {\color[HTML]{FF0000} 96.3}) & 
({\color[HTML]{0000EE} 94.6}, {\color[HTML]{0000EE} 98.4}) & 
({\color[HTML]{FF0000} 95.2}, {\color[HTML]{FF0000} 98.7}) & 
({\color[HTML]{FF0000} 97.5}, {\color[HTML]{FF0000} 99.0}) & 
({\color[HTML]{0000EE} 74.2}, {\color[HTML]{FF0000} 92.5}) & 
\textbf{({\color[HTML]{FF0000} 89.0}, {\color[HTML]{FF0000} 96.9})}
\\ \hline
& CLIP+R.            & (-, 17.9)    & (-, 68.5)     & (-, 89.5)    & (-, 4.7)     & (-, 74.3)     & (-, 22.1)     & (-, 47.5)     & (-, 82.7)     & (-, 73.6)     & (-, 78.9)     & (-, 56.0)                           \\
& PoinCLIP V2        & (40.6, 78.0) & (56.1, 84.4) & (53.8, 84.2)  & (52.7, 81.1)  & (50.7, 80.4)  & (40.8, 78.1)  & (54.9, 82.8)  & (48.9, 77.9)  & (54.3, 72.5)  & (59.3, 81.9)  & (78.3, 49.4)                       \\
& PointCLIP V2a      & (75.9, 40.8) & (76.2, 47.4) & (92.5, 79.9)  & (71.7, 30.7)  & (72.8, 44.9)  & (62.3, 21.9)  & (77.1, 46.4)  & (87.4, 63.7)  & (87.9, 69.9)  & (90.8, 70.8)  & (79.5, 51.6)                       \\
& AnomalyCLIP        & (93.7, 71.1)  & (90.7, 67.7)  & (95.8, 84.7)  & (82.0, 45.2)  & (93.9, 77.1)  & (84.3, 50.0)  & (93.5, 79.2)  & (95.6, 83.1)  & (95.9, 83.4)  & (91.2, 67.5)  & (91.6, 70.9)                        \\
& PointAD            & ({\color[HTML]{0000EE} 99.6}, {\color[HTML]{0000EE} 90.1}) & 
({\color[HTML]{0000EE} 96.7}, {\color[HTML]{FF0000} 97.9}) & 
({\color[HTML]{0000EE} 99.4}, {\color[HTML]{0000EE} 85.5}) & 
({\color[HTML]{FF0000} 92.6}, {\color[HTML]{0000EE} 85.4}) & 
({\color[HTML]{0000EE} 96.1}, {\color[HTML]{0000EE} 74.0}) & 
({\color[HTML]{0000EE} 92.4}, {\color[HTML]{FF0000} 98.3}) & 
({\color[HTML]{0000EE} 99.4}, {\color[HTML]{FF0000} 98.9}) & 
({\color[HTML]{0000EE} 99.8}, {\color[HTML]{0000EE} 92.9}) & 
({\color[HTML]{0000EE} 98.8}, {\color[HTML]{0000EE} 87.9}) & 
({\color[HTML]{0000EE} 97.5}, {\color[HTML]{FF0000} 91.1}) & 
({\color[HTML]{0000EE} (97.2}, {\color[HTML]{0000EE} 90.2}) 
 \\
% & CPMF (Un) & (96.6, -) & (92.8, -) & (95.5, -) & (92.8, -) & (92.8, -) & (89.0, -) & (79.0, -) & (98.7, -) & (97.2, -) & (96.8, -) & (96.9, -) \\
& MVP-WinCLIP    & (58.1, -)  & (61.4, -)  & (73.5, -)  & (71.8, -)  & (49.9, -)  & (50.2, -)  & (77.8, -)  & (71.1, -)  & (85.8, -) & (47.8, -)  & (64.7, -) \\
& MVP-PCLIP      & (97.1, -) & (82.2, -) & (94.4, -) & (84.7, -) & (72.0, -)  & (70.2, -) & (98.5, -) & (98.2, -) & (96.6, -) & (81.5, -)  & (87.5, -)        \\ \cline{2-13} 
\multirow{-8}{*}{\begin{tabular}[c]{@{}c@{}}Pixel-level\\ (AUROC, PRO)\end{tabular}} & MCL-AD            &
({\color[HTML]{FF0000} 99.6}, {\color[HTML]{FF0000} 98.8}) & 
({\color[HTML]{FF0000} 97.1}, {\color[HTML]{0000EE} 90.0}) & 
({\color[HTML]{FF0000} 99.5}, {\color[HTML]{FF0000} 97.8}) & 
({\color[HTML]{0000EE} 92.5}, {\color[HTML]{FF0000} 86.1}) & 
({\color[HTML]{FF0000} 96.2}, {\color[HTML]{FF0000} 86.3}) & 
({\color[HTML]{FF0000} 93.9}, {\color[HTML]{0000EE} 78.4}) & 
({\color[HTML]{FF0000} 99.4}, {\color[HTML]{0000EE} 98.2}) & 
({\color[HTML]{FF0000} 99.8}, {\color[HTML]{FF0000} 99.1}) & 
({\color[HTML]{FF0000} 99.1}, {\color[HTML]{FF0000} 94.6}) & 
({\color[HTML]{FF0000} 98.0}, {\color[HTML]{0000EE} 89.5}) & 
\textbf{({\color[HTML]{FF0000} 97.7}, {\color[HTML]{FF0000} 92.2})}
\\ \toprule[0.9pt]
\end{tabular}}}
\end{table*}
% \vspace{2cm}
\section{Experiments}
\subsection{Experimental Settings}
\textbf{Datasets.} This study primarily utilizes two datasets: MVTec3D-AD~\cite{ref20} and Eyecandies~\cite{ref33}. MVTec3D-AD is a widely used dataset for industrial 3D anomaly detection, includes 10 categories. It provides high-resolution point clouds and corresponding RGB images, captured using industrial 3D sensors and cameras. The Eyecandies dataset is a synthetic candy dataset consisting of 10 categories, which includes multimodal information such as RGB images, depth maps, and normal maps, along with precise anomaly annotations. In our ZS-3D anomaly detection task, the training sets of both datasets only consist of normal samples, so we exclusively use their test sets, which contain point-level annotations for point clouds and pixel-level annotations for images. We convert Eyecandies to a format compatible with MVTec3D-AD and project the point cloud data from both datasets into depth maps from 9 viewpoints for subsequent model training and evaluation.

\textbf{Evaluation Setting and Metrics.} To comprehensively evaluate the performance of our model in the ZS-3D anomaly detection, we employ two evaluation settings: within-dataset evaluation and cross-dataset evaluation. In the within-dataset setup, the model is evaluated on test objects that belong to different categories but come from the same dataset. In the cross-dataset setup, the model is trained on auxiliary objects from one dataset and directly tested on all objects from another dataset, thereby assessing the model's generalization ability under the condition of no target domain supervision. To systematically assess the model's performance in anomaly classification and localization, we used four commonly employed performance metrics: the Image-level area under the receiver operating characteristic curve (I-AUROC) and Average Precision (AP) to measure overall anomaly detection performance. Point-level AUROC (P-AUROC) and AUROC under restricted regions (AU-PRO) to evaluate the model's ability to localize anomalous regions at the pixel or point cloud level. All metric results are reported as percentages, with higher values indicating better performance.

\textbf{Implementation details.} The image encoder and text encoder are both based on CLIP~\cite{ref29}, with the input image size set to 336×336 pixels and an embedding feature dimension of 768. We select the $6_{th}$, $12_{th}$, $18_{th}$, and $24_{th}$ layers of the image encoder as key layers, and the learnable word embedding dimension is set to 14. The model is trained for 15 epochs with a learning rate of 0.001. All model inference is conducted on a single NVIDIA RTX 4090 24GB GPU. We generate 9 views using the Open3d library by rotating the point cloud along the X and Y axis with angles of \{$-\frac{\pi}{4}$, $-\frac{\pi}{12}$, $0$, $\frac{\pi}{4}$, $\frac{\pi}{12}$\} and \{$-\frac{\pi}{4}$, $-\frac{\pi}{12}$, $\frac{\pi}{4}$, $\frac{\pi}{12}$\}, which is suitable for most categories.

\subsection{Comparison with State-of-the-Arts}
To comprehensively evaluate the performance of ZA-3D anomaly detection, we compares the proposed MCL-AD with several state-of-the-art baseline methods. These models include MVP-Winclip, MVP-SAA, MVP-April-Gan, CLIP + R, PointCLIP V2~\cite{ref45}, AnomalyCLIP~\cite{ref46}, PointAD~\cite{ref31}, and MVP-PCLIP~\cite{ref27}. MVP-Winclip is developed by integrating the zero-shot image anomaly detection method Winclip~\cite{ref9} with the MVP framework. CLIP + R combines the original CLIP with rendering techniques for 3D anomaly detection. PointCLIP V2 performs zero-shot 3D classification by projecting point clouds into depth maps, while AnomalyCLIP adapts a 2D anomaly detection method for 3D anomaly detection through fine-tuning. PointAD and MVP-PCLIP are specifically designed for zero-shot 3D anomaly detection in point clouds.

\subsubsection{Quantitative Comparison}
\textbf{On the MVTec3D-AD Dataset.} As shown in Tab.~\ref{tab:1}, MVP-PCLIP achieves a limited I-AUROC of 71.9\% due to its reliance on point cloud geometry via multi-view CLIP projections, which lack the texture information needed for complex scenarios. Similarly, CLIP+R performs poorly in anomaly localization tasks, achieving an AUPRO of only 56.0\%, due to its focus on object-level semantic representations and limited capacity to discriminate fine-grained local anomalies. In contrast, our proposed MCL-AD significantly improves both anomaly detection and localization performance. Compared to MVP-PCLIP, MCL-AD improves I-AUROC by 17\% and AP by 7.3\%. Moreover, it surpasses CLIP+R in localization accuracy, with AUPRO increased by a remarkable 36.1\%. These improvements are attributed to our Multimodal Prompt Learning Mechanism (MPLM) and Collaborative Modulation Mechanism (CMM), which effectively integrates the complementary information from RGB images and point cloud data.

\begin{table*}[t!]
\setlength{\tabcolsep}{2pt}
\renewcommand{\arraystretch}{1.5}
\caption{\textrm{Quantitative results on the Eyecandies dataset (\%).}}
\label{tab:2}
\scriptsize
\centering
\resizebox{\linewidth}{!}{
\rmfamily
\begin{tabular}{clccccccccccc}
\toprule[0.9pt]
\multicolumn{1}{l}{}                                                                & Method        & \begin{tabular}[c]{@{}c@{}}Candy \\ Cane\end{tabular} & \begin{tabular}[c]{@{}c@{}}Chocolate \\ Cookie\end{tabular} & \begin{tabular}[c]{@{}c@{}}Chocolate \\ Praline\end{tabular} & Confetto     & \begin{tabular}[c]{@{}c@{}}Gummy \\ Bear\end{tabular} & \begin{tabular}[c]{@{}c@{}}Hazelnut \\ Truffle\end{tabular} & \begin{tabular}[c]{@{}c@{}}Licorice \\ Sandwich\end{tabular} & Lollipop     & Marshmallow  & \begin{tabular}[c]{@{}c@{}}Peppermint \\ Candy\end{tabular} & Mean         \\ \hline
\multirow{7}{*}{\begin{tabular}[c]{@{}c@{}}Image-level\\ (AUROC, AP)\end{tabular}}  & CLIP+R.       & (64.3, 67.8)     & (76.6, 77.6)  & (64.3, 70.8)  & (88.0, 89.6) & (70.4, 72.1)   & (55.5, 53.8)   & (78.4, 81.9) & (71.5, 64.7) & (77.1, 77.8) & (83.4, 82.7)   & (73.0, 73.9) \\
& PoinCLIP V2   & (43.0, 48.3)     & (48.0, 55.0)     & (46.4, 51.6)   & (49.3, 48.4) & (44.7, 49.1)   & (48.3, 55.4)  & (61.8, 70.0)    & (42.1, 30.4)  & (54.1, 51.5)  & (31.2, 39.6)                                                 & (46.9, 49.9) \\
 & PointCLIP V2a & (44.1, 51.2)    & (44.5, 52.4)   & (48.6, 52.3)    & (56.7, 54.8) & (42.8, 44.6)   & (55.7, 60.3)       & (63.5, 68.7)    & (43.8, 29.3)  & (54.0, 52.1)  & (31.3, 39.6)     & (48.5, 50.5)  \\
& AnomalyCLIP   & (49.7, 50.8)    & (57.1, 62.9)    & (66.5, 70.3)      & (66.7, 68.0) & (64.0, 69.5)   & (61.1, 67.8)      & (69.2, 73.9)   & (68.8, 56.5) & (77.3, 80.4)  & (69.5, 74.6)       & (65.0, 67.5)  \\
 & PointAD       & ({\color[HTML]{FF0000} 42.8}, {\color[HTML]{FF0000} 49.1}) & ({\color[HTML]{0000EE} 85.3}, {\color[HTML]{0000EE} 89.3}) & ({\color[HTML]{0000EE} 86.6}, {\color[HTML]{0000EE} 89.8}) & ({\color[HTML]{0000EE} 89.5}, {\color[HTML]{0000EE} 92.3}) & ({\color[HTML]{0000EE} 75.3}, {\color[HTML]{0000EE} 77.8}) & ({\color[HTML]{0000EE} 61.4}, {\color[HTML]{0000EE} 68.5}) & ({\color[HTML]{0000EE} 87.2}, {\color[HTML]{0000EE} 89.0}) & ({\color[HTML]{0000EE} 70.4}, {\color[HTML]{0000EE} 63.9}) & ({\color[HTML]{0000EE} 86.8}, {\color[HTML]{FF0000} 89.5}) & ({\color[HTML]{FF0000} 91.8}, {\color[HTML]{FF0000} 94.2}) & ({\color[HTML]{0000EE} 77.7}, {\color[HTML]{0000EE} 80.4})  \\ \cline{2-13} 
& MCL-AD       & ({\color[HTML]{0000EE} 42.5}, {\color[HTML]{0000EE} 46.9}) & ({\color[HTML]{FF0000} 85.6}, {\color[HTML]{FF0000} 90.3}) & ({\color[HTML]{FF0000} 91.0}, {\color[HTML]{FF0000} 92.5}) & ({\color[HTML]{FF0000} 93.4}, {\color[HTML]{FF0000} 95.8}) & ({\color[HTML]{FF0000} 80.6}, {\color[HTML]{FF0000} 83.5}) & ({\color[HTML]{FF0000} 70.4}, {\color[HTML]{FF0000} 72.3}) & ({\color[HTML]{FF0000} 89.0}, {\color[HTML]{FF0000} 91.0}) & ({\color[HTML]{FF0000} 73.7}, {\color[HTML]{FF0000} 67.1}) & ({\color[HTML]{FF0000} 79.8}, {\color[HTML]{0000EE} 83.5}) & ({\color[HTML]{0000EE} 89.4}, {\color[HTML]{0000EE} 93.8}) & \textbf{({\color[HTML]{FF0000} 78.3}, {\color[HTML]{FF0000} 80.8})} \\ \hline
\multirow{7}{*}{\begin{tabular}[c]{@{}c@{}}Pixel-level\\ (AUROC, PRO)\end{tabular}} & CLIP+R.       & (97.3, 89.2)       & (72.8, 10.3)    & (65.5, 8.3)    & (75.0, 17.4) & (83.9, 31.8)    & (72.4, 29.8)    & (75.6, 17.4)   & (94.4, 75.8) & (66.3, 15.7) & (76.7,22.2)   & (78.0, 31.8) \\
  & PoinCLIP V2   & (44.8, -)     & (44.8, -)     & (48.0, -)       & (59.6, -)     & (48.6, -)    & (53.9, -)          & (42.2, -)    & (33.7, -)     & (43.3, -)     & (41.4, -)    & (46.0, -)     \\
 & PointCLIP V2a & (44.8, -)     & (44.7, -)      & (49.0, -)      & (59.3, -)     & (48.2, -)   & (54.2, -)        & (42.2, -)      & (33.6, -)     & (45.0,-)     & (41.5, -)      & (46.2, -)     \\
  & AnomalyCLIP   & (96.7, 88.5)     & (89.5, 64.3)      & (83.9, 55.4)   & (92.1, 74.8) & (77.4, 33.1)   & (70.2, 27.3)    & (83.0, 52.9)    & (94.6, 78.9) & (77.9, 38.2) & (84.8, 48.8)                                                & (85.0, 56.2) \\
 & PointAD       & ({\color[HTML]{0000EE} 96.4}, {\color[HTML]{0000EE} 87.2}) & ({\color[HTML]{0000EE} 97.8}, {\color[HTML]{0000EE} 90.3}) & ({\color[HTML]{0000EE} 93.5}, {\color[HTML]{0000EE} 83.7}) & ({\color[HTML]{0000EE} 98.7}, {\color[HTML]{0000EE} 94.7}) & ({\color[HTML]{0000EE} 90.8}, {\color[HTML]{0000EE} 74.4}) & ({\color[HTML]{0000EE} 87.7}, {\color[HTML]{0000EE} 62.5}) & ({\color[HTML]{0000EE} 96.8}, {\color[HTML]{0000EE} 88.0}) & ({\color[HTML]{0000EE} 96.7}, {\color[HTML]{0000EE} 85.4}) & ({\color[HTML]{0000EE} 97.6}, {\color[HTML]{FF0000} 88.5}) & ({\color[HTML]{0000EE} 97.1}, {\color[HTML]{0000EE} 88.6}) & ({\color[HTML]{0000EE} 95.3}, {\color[HTML]{0000EE} 84.3})    \\ \cline{2-13} 
 & MCL-AD       
 & ({\color[HTML]{FF0000} 97.1}, {\color[HTML]{FF0000} 89.4})                                                   &  ({\color[HTML]{FF0000} 98.1}, {\color[HTML]{FF0000} 91.3})                                                   & ({\color[HTML]{FF0000} 95.1}, {\color[HTML]{FF0000} 87.1})                                                   &  ({\color[HTML]{FF0000} 98.9}, {\color[HTML]{FF0000} 95.6})             
 &  ({\color[HTML]{FF0000} 92.5}, {\color[HTML]{FF0000} 79.1})                                                   &  ({\color[HTML]{FF0000} 91.8}, {\color[HTML]{FF0000} 70.7})                                                   &     ({\color[HTML]{FF0000} 97.6}, {\color[HTML]{FF0000} 89.2})                                          &   ({\color[HTML]{FF0000} 97.5}, {\color[HTML]{FF0000} 84.9})            
 &    ({\color[HTML]{FF0000} 97.4}, {\color[HTML]{0000EE} 86.7})          
 & ({\color[HTML]{FF0000} 98.2}, {\color[HTML]{FF0000} 90.1})                                                   & \textbf{({\color[HTML]{FF0000} 96.3}, {\color[HTML]{FF0000} 86.0})} \\ \toprule[0.9pt]
\end{tabular}}
\end{table*}
%Performance comparison 
\begin{table*}[t!]
\setlength{\tabcolsep}{2pt}
\renewcommand{\arraystretch}{1.5}
\caption{\textrm{Quantitative results on MCL-AD cross-dataset anomaly detection transferring from MVTec3D-AD to Eyecandies.}}
\label{tab:3}
\scriptsize
\centering
\resizebox{\linewidth}{!}{
\rmfamily
\begin{tabular}{cllllllllllll}
\toprule[0.9pt]
\multicolumn{1}{l}{}  & Method & \multicolumn{1}{c}{\begin{tabular}[c]{@{}c@{}}Candy \\ Cane\end{tabular}} & \multicolumn{1}{c}{\begin{tabular}[c]{@{}c@{}}Chocolate \\ Cookie\end{tabular}} & \multicolumn{1}{c}{\begin{tabular}[c]{@{}c@{}}Chocolate \\ Praline\end{tabular}} & \multicolumn{1}{c}{Confetto} & \multicolumn{1}{c}{{\color[HTML]{1F2328} \begin{tabular}[c]{@{}c@{}}Gummy \\ Bear\end{tabular}}} & \multicolumn{1}{c}{{\color[HTML]{1F2328} \begin{tabular}[c]{@{}c@{}}Hazelnut \\ Truffle\end{tabular}}} & \multicolumn{1}{c}{\begin{tabular}[c]{@{}c@{}}Licorice \\ Sandwich\end{tabular}} & \multicolumn{1}{c}{Lollipop} & \multicolumn{1}{c}{Marshmallow} & \multicolumn{1}{c}{\begin{tabular}[c]{@{}c@{}}Peppermint \\ Candy\end{tabular}} & \multicolumn{1}{c}{Mean} \\ \hline & PoinCLIP V2  & (42.1, 50.4)  & (45.5, 54.9)   & (49.0, 52.5)   & (57.1, 54.3)   & (44.9, 45.8)  & (53.6, 56.6)  & (59.0, 64.5) & (47.4, 36.1)                 & (54.1, 53.3)   & (32.6, 40.3)   & (48.5, 50.9)   \\
& AnomalyCLIP  & (44.9, 51.4)  & (66.1, 68.9)   & (76.4, 77.7)          & (79.3, 82.7)  & (51.0, 56.7)  & (55.1,58.3)  & (75.2, 79.6)   & (65.4, 57.9) & (70.3, 72.5) & (72.8, 75.5)  & (65.7, 68.1)        \\
& PointAD   & ({\color[HTML]{FF0000} 49.4}, {\color[HTML]{FF0000} 55.1}) & ({\color[HTML]{0000EE} 87.4}, {\color[HTML]{0000EE} 91.2}) & ({\color[HTML]{0000EE} 87.5}, {\color[HTML]{0000EE} 88.8}) & ({\color[HTML]{0000EE} 91.0}, {\color[HTML]{0000EE} 94.1}) & ({\color[HTML]{0000EE} 71.4}, {\color[HTML]{0000EE} 74.2}) & ({\color[HTML]{0000EE} 70.2}, {\color[HTML]{0000EE} 75.0}) & ({\color[HTML]{FF0000} 88.0}, {\color[HTML]{0000EE} 89.5}) & ({\color[HTML]{0000EE} 73.2}, {\color[HTML]{0000EE} 63.5}) & ({\color[HTML]{0000EE} 79.4}, {\color[HTML]{0000EE} 84.4}) & ({\color[HTML]{0000EE} 88.2}, {\color[HTML]{0000EE} 91.9}) & ({\color[HTML]{0000EE} 78.6}, {\color[HTML]{0000EE} 80.8})       \\ \cline{2-13} 
\multirow{-4}{*}{\begin{tabular}[c]{@{}c@{}}Image-level\\ (AUROC, AP)\end{tabular}}  & MCL-AD         & ({\color[HTML]{0000EE} 47.5}, {\color[HTML]{0000EE} 49.5}) & 
({\color[HTML]{FF0000} 89.4}, {\color[HTML]{FF0000} 92.8}) & 
({\color[HTML]{FF0000} 90.9}, {\color[HTML]{FF0000} 93.2}) & 
({\color[HTML]{FF0000} 92.7}, {\color[HTML]{FF0000} 94.7}) & 
({\color[HTML]{FF0000} 75.6}, {\color[HTML]{FF0000} 79.8}) & 
({\color[HTML]{FF0000} 76.7}, {\color[HTML]{FF0000} 81.5}) & 
({\color[HTML]{0000EE} 87.9}, {\color[HTML]{FF0000} 91.5}) & 
({\color[HTML]{FF0000} 78.6}, {\color[HTML]{FF0000} 71.0}) & 
({\color[HTML]{FF0000} 81.0}, {\color[HTML]{FF0000} 85.1}) & 
({\color[HTML]{FF0000} 90.4}, {\color[HTML]{FF0000} 93.2}) & 
\textbf{({\color[HTML]{FF0000} 81.1}, {\color[HTML]{FF0000} 83.2})}
  \\ \hline

& PoinCLIP V2   & (45.1, -)   & (47.2, -)  & (48.1, -)  & (63.1, -)  & (49.7, -)   & (54.0, 15.6)   & (45.2, -)    & (33.8, -)                     & (44.7, -)   & (41.9, -)    & (47.3, -)   \\
 & AnomalyCLIP   & (95.1, 82.3)   & (91.6, 71.2)   & (84.3, 62.0)      & (94.1, 80.9)  & (80.6, 44.5)  & (73.6, 42.7)  & (89.6, 66.9)  & (92.6, 68.0)  & (75.9, 42.8)  & (84.9, 52.1) & (94.4, 80.3) \\
& PointAD & ({\color[HTML]{0000EE} 88.8}, {\color[HTML]{0000EE} 63.9}) & ({\color[HTML]{FF0000} 97.4}, {\color[HTML]{FF0000} 88.9}) & ({\color[HTML]{0000EE} 92.5}, {\color[HTML]{0000EE} 85.1}) & ({\color[HTML]{FF0000} 99.1}, {\color[HTML]{FF0000} 96.6}) & ({\color[HTML]{0000EE} 89.9}, {\color[HTML]{0000EE} 70.4}) & ({\color[HTML]{0000EE} 89.9}, {\color[HTML]{0000EE} 70.4}) & ({\color[HTML]{0000EE} 95.5}, {\color[HTML]{0000EE} 84.4}) & ({\color[HTML]{0000EE} 95.7}, {\color[HTML]{0000EE} 79.8}) & ({\color[HTML]{0000EE} 95.8}, {\color[HTML]{0000EE} 82.5}) & ({\color[HTML]{0000EE} 95.2}, {\color[HTML]{0000EE} 84.8}) & ({\color[HTML]{0000EE} 94.0}, {\color[HTML]{0000EE} 80.7}) \\ \cline{2-13} 
\multirow{-4}{*}{\begin{tabular}[c]{@{}c@{}}Pixel-level\\ (AUROC, PRO)\end{tabular}} & MCL-AD            & ({\color[HTML]{FF0000} 98.1}, {\color[HTML]{FF0000} 91.5}) & 
({\color[HTML]{0000EE} 94.2}, {\color[HTML]{0000EE} 87.4}) & 
({\color[HTML]{FF0000} 99.2}, {\color[HTML]{FF0000} 97.0}) & 
({\color[HTML]{0000EE} 92.1}, {\color[HTML]{0000EE} 78.0}) & 
({\color[HTML]{FF0000} 94.0}, {\color[HTML]{FF0000} 77.5}) & 
({\color[HTML]{FF0000} 97.3}, {\color[HTML]{FF0000} 90.0}) & 
({\color[HTML]{FF0000} 96.9}, {\color[HTML]{FF0000} 87.3}) & 
({\color[HTML]{FF0000} 97.3}, {\color[HTML]{FF0000} 86.5}) & 
({\color[HTML]{FF0000} 95.8}, {\color[HTML]{FF0000} 85.4}) & 
({\color[HTML]{FF0000} 96.9}, {\color[HTML]{FF0000} 88.2}) & 
\textbf{({\color[HTML]{FF0000} 95.8}, {\color[HTML]{FF0000} 85.4})}
 \\		
\bottomrule
\end{tabular}}
\end{table*}

\textbf{On the Eyecandies Dataset.} As shown in Tab.~\ref{tab:2}, MCL-AD outperforms existing methods on the Eyecandies dataset. Compared to the projection-based PointCLIP V2, MCL-AD demonstrates superior ZS-3D anomaly detection performance, with I-AUROC improving from 46.9\% to 78.3\% and AP increasing from 49.9\% to 80.8\%. Additionally, MCL-AD shows significant improvements in ZS-3D anomaly localization, with P-AUROC increasing by 95.3\% to 96.3\% compared to PointAD, and AUPRO rising from 84.3\% to 86.0\%. These performance gains are attributed to MCL-AD's ability to effectively leverage complementary information from RGB images and point clouds, enhancing the understanding and integration of different modal information through collaborative modulation and multimodal prompt learning mechanisms.

\textbf{On the Cross-Dataset.} 
To further evaluate the zero-shot capability of MCL-AD, we conduct cross-dataset anomaly detection experiments, as shown in Tab.~\ref{tab:3}. In this setting, the model is tested on objects with entirely different semantics and scenes, using objects from one dataset as auxiliary references, thereby assessing its generalization ability. Specifically, AnomalyCLIP relies on object-agnostic prompts and a specialized architecture, which limits its adaptability. In contrast, MCL-AD achieves superior cross-dataset performance by leveraging the interaction between RGB image and point cloud prompts, fully exploiting the complementary strengths of the two modalities. Concretely, MCL-AD outperforms AnomalyCLIP by 15.4\% and 15.1\% on image-level and pixel-level metrics, respectively, and further achieves 1.4\% and 5.1\% improvements on the I-AUROC and P-AUROC scores. These results demonstrate that MCL-AD exhibits strong generalization capabilities in unseen zero-shot anomaly detection tasks.

\begin{figure*}[t!]
    \centering
\includegraphics[width=1\linewidth]{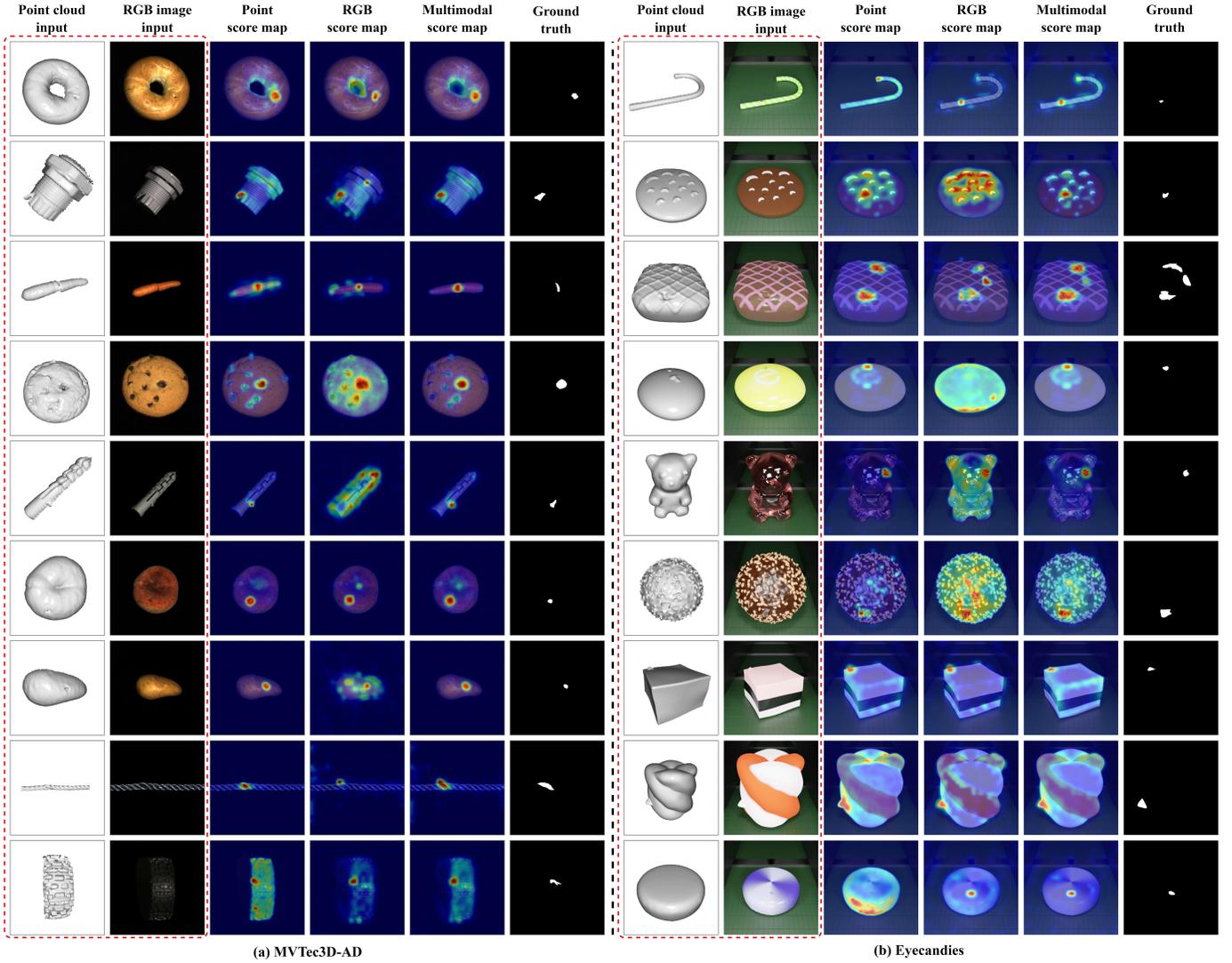}
    \caption{\textrm{Visualization results on the MVTec3D-AD and Eyecandies datasets.}}
    \label{fig:4}
\end{figure*}

\begin{figure}[t!]
    \centering
\includegraphics[width=1\linewidth]{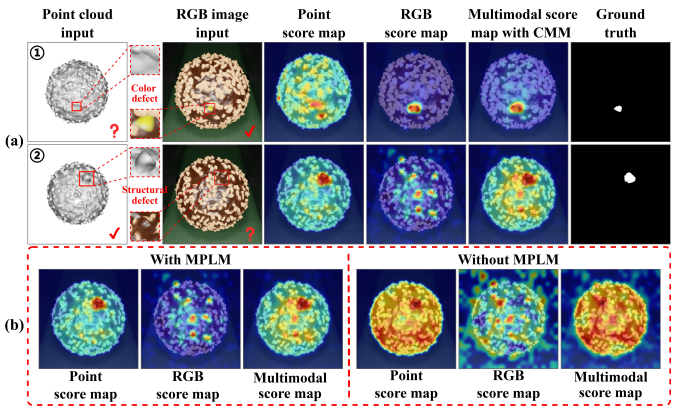}
    \caption{\textrm{Visualization of the effectiveness of the key module.}}
    \label{fig:5}
\end{figure}

\subsubsection{Qualitative Analysis} To visually demonstrate the model's superiority in detecting unseen anomalous situations, we present the point clouds and corresponding RGB images, along with the multimodal anomaly score maps of MCL-AD on the MVTec3D-AD and Eyecandies datasets. As shown in Fig.~\ref{fig:4}, it is evident that MCL-AD exhibits exceptional anomaly detection and localization capabilities on both point clouds and RGB images, with a significant improvement in detection performance through the multimodal collaboration learning.

\textbf{Visualization of the effectiveness of the key module.} As shown in Fig.~\ref{fig:5}(a), we further visualize the effectiveness of the MPLM and CMM. The first row displays the image of ``HazelnutTruffle'', where color contamination is present but no geometric anomaly exists, making it difficult to detect such anomalies using only point cloud data. In contrast, the second row's ``HazelnutTruffle'' sample exhibits a clear structural anomaly that, despite having a similar color to the object foreground, can be distinctly identified through point cloud information. As shown in multimodal score map MCL-AD effectively integrates both modalities through the CMM, complementing their respective strengths. We further investigate the effectiveness of MPLM. As shown in Fig.~\ref{fig:5}(b), when the MPLM is not applied, noise increases significantly, leading to suboptimal multimodal representations. This underscores the importance of the collaborative optimization of point cloud and RGB information. Therefore, we conclude that the collaborative optimization of point cloud and RGB image data during training endows the model with powerful multimodal detection capabilities.

\begin{table*}[ht!]
\setlength{\tabcolsep}{14pt}
\renewcommand{\arraystretch}{1.2}
\caption{\textrm{Ablation on the proposed modules.}}
\label{tab:4}
\scriptsize
\centering
\resizebox{\linewidth}{!}{
\rmfamily
\begin{tabular}{ccccccccc}
\toprule[0.8pt]
\multicolumn{3}{c}{} & \multicolumn{2}{c}{MPLM} & \multicolumn{2}{c}{Image-level}                      & \multicolumn{2}{c}{Pixel-level}                         \\ 
\cmidrule(lr){1-3}
\cmidrule(lr){4-5}
\cmidrule(lr){6-7}
\cmidrule(lr){8-9}
$L_{point}$   & $L_{rgb}$  & CMM  & ODTP  & $L_{mcl}$  & \multicolumn{1}{l}{I-AUROC} & \multicolumn{1}{l}{AP} & \multicolumn{1}{l}{P-AUROC} & \multicolumn{1}{l}{AUPRO} \\ \hline
         &       &      &                &                     & 60.4                        & 86.4     & -                           & 56.0               
\\
\checkmark       &  &     &      &    &78.8       &93.3  &92.2   & 80.4                 
\\
\checkmark        & \checkmark     &      &                &          &  85.3        &  95.5                           &  96.8                      &  89.0              
\\
\checkmark       & \checkmark     & \checkmark    &                &         & 86.8                            &96.1                        & 97.0                            &89.6    
\\   \checkmark
& \checkmark       & \checkmark      & \checkmark         &     &  88.3       &  96.7                                                   &  97.3                           &   91.1  
                        
\\
\checkmark        & \checkmark     & \checkmark    & \checkmark              & \checkmark      & \textbf{89.0} & \textbf{96.9} & \textbf{97.7} & \textbf{92.2}                        \\ \toprule[0.8pt]
\end{tabular}}
\end{table*}

\begin{figure*}[ht!]
    \centering
    \includegraphics[width=0.8\textwidth]{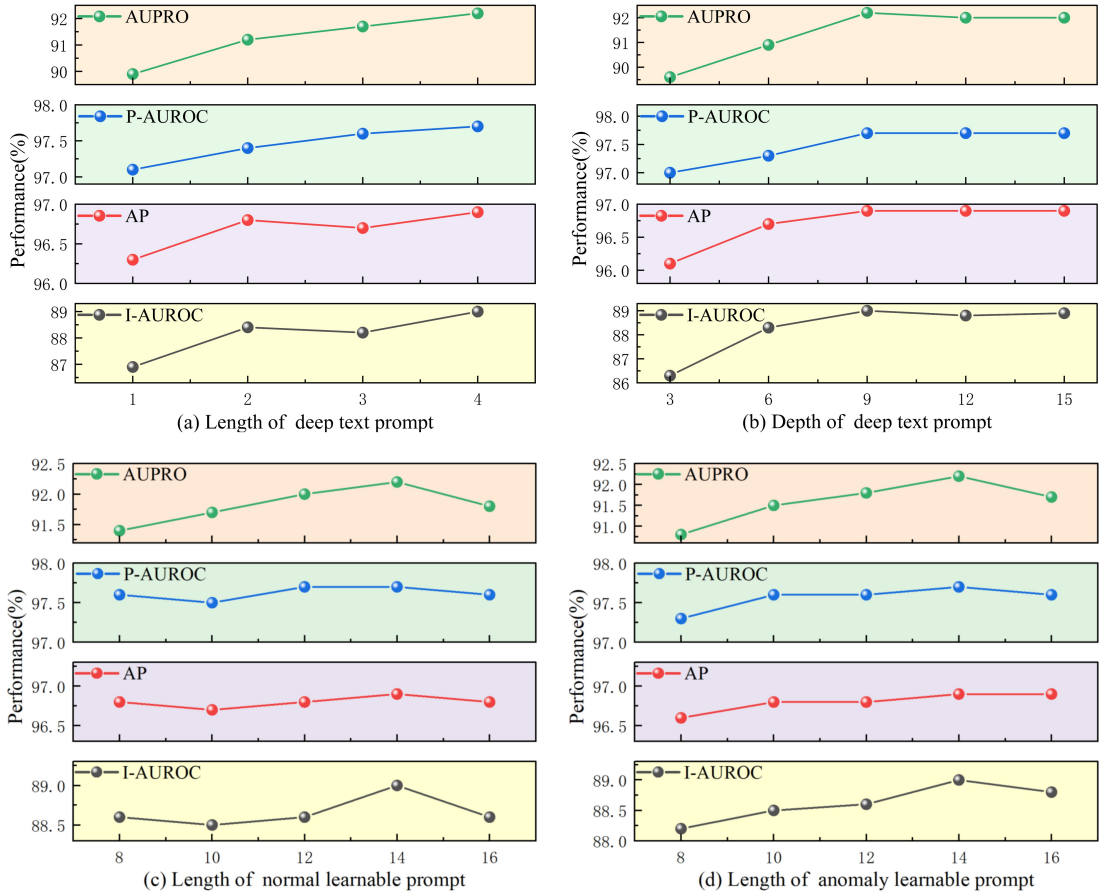}
    \caption{\textrm{Ablation Study on Textual Prompt Hyperparameters. (a) is the length of normal learnable prompt, (b) is the length of anomaly learnable prompt, (c) is the length of deep-text prompts, and (d) is the depth of deep-text prompts.}}
    \label{fig:3}
\end{figure*}

\subsection{Ablation Study}

\subsubsection{Key Modules of MCL-AD} To validate the effectiveness of the proposed modules, we adopt a stepwise integration strategy for evaluation. As shown in Table~\ref{tab:4}, the CLIP-based rendering method demonstrates limited performance in both image-level and pixel-level anomaly detection tasks. This limitation arises because CLIP primarily focuses on 2D object-level semantic alignment and struggles to model anomalous features. To enhance the modeling of anomalous semantics, we introduce point cloud anomaly information ($L_{point}$), which significantly improves detection performance. However, due to the point cloud modality's lower sensitivity to texture anomalies such as color, the model still faces challenges in capturing fine-grained anomalies in complex scenes. To address this issue, we further introduce the $L_{rgb}$ loss to enhance the model's perception of texture anomalies, thereby improving image-level detection performance. Traditional multimodal fusion methods rely on direct fusion, resulting in inefficient inter-modal interactions. To this end, MCL-AD proposes a Collaborative Modulation Mechanism (CMM), which significantly improves fusion quality, raising the I-AUROC from 85.3\% to 86.8\%. Additionally, we design a Object-agnostic Decoupled Text Prompt (ODTP) to more effectively learn anomaly representations in point clouds and RGB images. By introducing a multimodal contrastive loss ($L_{mcl}$), we further strengthen the semantic associations between modalities, significantly enhancing multimodal segmentation performance, with the P-AUROC improving from 60.4\% to 97.7\% and the AUROC from 86.4\% to 92.0\%.

\begin{table}[ht!]
\setlength{\tabcolsep}{8pt}
\renewcommand{\arraystretch}{1.6}
\caption{\textrm{Ablation on anchor prompt.}}
\label{tab:5}
\centering
\resizebox{\linewidth}{!}{
\rmfamily
\begin{tabular}{lcccc}
\toprule[0.8pt]
Anchor prompt & I-AUROC & AP & P-AUROC & AUPRO \\ \hline
Point prompt  & \textbf{89.0} & \textbf{96.9} & \textbf{97.7} & \textbf{92.2}  \\
RGB prompt    &  88.6  & 96.8  &  97.6   &   91.9 
    \\ \toprule[0.8pt]
\end{tabular}}
\end{table}

\subsubsection{Anchor Setup of MCL}
To demonstrate the effectiveness of the anchor prompt design in MCL, we conducted the ablation study shown in Tab.~\ref{tab:5}. The results clearly indicate that point cloud prompts consistently outperform RGB prompts across all domains and evaluation metrics. The study suggests that using point cloud prompts as anchor points provides a more comprehensive feature representation, thereby enhancing overall performance.
\begin{table}[h!]
\setlength{\tabcolsep}{8pt}
\renewcommand{\arraystretch}{1.2}
\caption{\textrm{Ablation on normal and anomaly prompt positions.}}
\label{tab:6}
\centering
\resizebox{\linewidth}{!}{
\rmfamily
\begin{tabular}{ccccc}
\toprule[0.8pt]
Prompt position                     & I-AUROC              & AP                   & P-AUROC              & AUPRO                \\ \hline
{[}P/R{]}{[}N/A{]}{[}object{]} & \textbf{89.0} & \textbf{96.9} & \textbf{97.7} & \textbf{92.2}                     \\
{[}P/R{]}{[}object{]}{[}N/A{]} &  88.4                    & 96.7                     &   97.6                   &  91.9                    \\
{[}N/A{]}{[}P/R{]}{[}object{]}&  87.9                    & 96.6                     &   97.3                  &  90.8    \\ \toprule[0.8pt]
\end{tabular}}
\end{table}

\subsubsection{Positioning of Decoupled Prompts} We conduct experiments to evaluate the impact of the positioning of learnable prompt tokens in decoupled prompts on performance. As shown in Tab.~\ref{tab:6}, placing the modality identifier at the beginning of the sentence and positioning the learnable prompt token in the middle (e.g., [P/R][N/A][object]) significantly improves performance. This arrangement enhances model guidance by establishing the topic at the beginning of the sentence, thereby exerting a stronger influence on the model.

\subsubsection{Decoupled Semantic Text Prompt Hyperparameter} We systematically analyze the impact of several key hyperparameters in decoupled textual prompts on model performance. As shown in Fig.~\ref{fig:3}(a), the model achieves optimal performance when the length of  deep text prompt  is set to 4. Fig.~\ref{fig:3}(b) illustrates the effect of prompt depth within the text encoder, showing that inserting prompts at the 9th layer yields the best performance, while shallower depths result in performance degradation. Fig.~\ref{fig:3}(c) and Fig.~\ref{fig:3}(d) further demonstrate that both normal and anomaly prompts perform best when the learnable prompt length is set to 14 tokens. Overall, an excessive or insufficient number of embedded prompt tokens may compromise the model’s performance.
\begin{table}[h!]
\setlength{\tabcolsep}{12pt}
\renewcommand{\arraystretch}{1}
\caption{\textrm{Ablation on the impact of different hyper-parameters $\eta$ on model performance.}}
\label{tab:7}
\centering
\resizebox{\linewidth}{!}{
\rmfamily
\begin{tabular}{ccccc}
\toprule[0.8pt]
 $\eta$ & \multicolumn{1}{c}{I-AUROC} & \multicolumn{1}{c}{AP} & \multicolumn{1}{c}{P-AUROC} & \multicolumn{1}{c}{AUPRO} \\ \hline
0.6      & \multicolumn{1}{c}{88.7}        & \multicolumn{1}{c}{96.8}   & \multicolumn{1}{c}{97.6}        & \multicolumn{1}{c}{92.0}      \\
0.8   & \textbf{89.0} & \textbf{96.9} & \textbf{97.7} & \textbf{92.2}       \\
1.0   &  87.9                           &96.5                        & 97.6                            &  91.8                         \\
1.2   & 87.4                            & 96.4                       &  97.5                           & 91.6              \\
1.4      &86.1                             & 96.0                       & 97.4                            & 91.1                          \\ 
\toprule[0.8pt]
\end{tabular}}
\end{table}

\subsubsection{Collaborative Modulation Mechanism} 
\label{sec:4.4}
Tab.~\ref{tab:7} reports the performance of the CMM on the MVTec3D-AD dataset. The fusion coefficient $\eta$ controls the weighting between the point cloud-guided branch and the RGB image-guided branch. As shown in Tab.~\ref{tab:7}, the model achieves the best performance when $\eta$ = 0.8, suggesting that a fusion scheme primarily driven by point cloud features, with RGB features playing a supplementary role, is better aligned with the characteristics of this dataset. The results reveal notable inter-domain differences in anomaly distributions, making static fusion strategies insufficient for diverse scenarios. To overcome this, we propose the CMM, which dynamically adjusts branch fusion to improve adaptability and robustness in anomaly detection and segmentation.

\begin{table}[h!]
\setlength{\tabcolsep}{12pt}
\renewcommand{\arraystretch}{1}
\caption{\textrm{Ablation on the impact of different hyper-parameters $\lambda_3$ on model performance.}}
\label{tab:8}
\centering
\resizebox{\linewidth}{!}{
\rmfamily
\begin{tabular}{ccccc}
\toprule[0.8pt]
$\lambda_3$ & \multicolumn{1}{c}{I-AUROC} & \multicolumn{1}{c}{AP} & \multicolumn{1}{c}{P-AUROC} & \multicolumn{1}{c}{AUPRO} \\ \hline
0.01      & \multicolumn{1}{c}{88.3}        & \multicolumn{1}{c}{96.7}   & \multicolumn{1}{c}{97.3}        & \multicolumn{1}{c}{91.1}      \\
0.1  & \multicolumn{1}{c}{88.4}        & \multicolumn{1}{c}{96.7}   & \multicolumn{1}{c}{97.5}        & \multicolumn{1}{c}{91.4}      \\
0.6   &  88.7                           & 96.8                       &  97.6                           &  91.4                         \\
0.8    & \textbf{89.0} & \textbf{96.9} & \textbf{97.7} & \textbf{92.2}                         \\
1      &88.6                             & 96.8                       &  97.6                           &  91.6                         \\ \toprule[0.8pt]
\end{tabular}}
\end{table}
\subsubsection{Multimodal Contrastive Learning}
\label{sec:4.5}
In this study, we systematically evaluate the impact of the hyperparameter $\lambda_3$ on the performance trade-off between point cloud and RGB representations in the ZSAD model. To achieve complementary learning of the two modalities, we introduce a multimodal contrastive learning approach. Notably, even with a decoupled design of point cloud and RGB prompts, zero-shot anomaly detection already outperforms existing state-of-the-art methods (see the first row of Tab.~\ref{tab:8}). Building upon this, the incorporation of a multimodal contrastive loss further enhances the model's generalization capability, with the best performance observed when $\lambda_3$ is set to 0.8.

\section{Conclusion}
In this paper, we proposed a MCL-AD that leverages multimodal collaboration learning among point clouds, RGB images, and textual semantics to address the challenges of ZS-3D anomaly detection. Firstly, we introduced two key components: the Multimodal Prompt Learning Mechanism (MPLM) and the Collaborative Modulation Mechanism (CMM). MPLM enhances the intra- and inter-modal representational capacity by introducing an object-agnostic decoupled semantic text prompt and a multimodal contrastive loss, enabling the model to learn more discriminative and synergistic features across modalities. CMM further improves inference robustness by jointly modulating RGB-guided and point cloud-guided branches, effectively addressing modality imbalance through a dual-modality comparison architecture. Extensive experiments on standard benchmarks, including MVTec3D-AD and Eyecandies, validate the effectiveness of MCL-AD, demonstrating superior performance and generalizability compared to existing state-of-the-art zero-shot 3D anomaly detection methods. In the future, we plan to explore extending MCL-AD to open-set scenarios and real-world industrial applications.

\bibliographystyle{IEEEtran}
\bibliography{ref}

\vfill

\end{document}